\documentclass[10pt,twocolumn,letterpaper]{article}

\usepackage[pagenumbers]{iccv}              % To produce the CAMERA-READY version

\usepackage{url}
\usepackage{graphicx}
\usepackage{xspace}
\usepackage{colortbl}
\usepackage{booktabs}
\usepackage{multirow,marvosym}

\usepackage{xcolor}

\definecolor{ForestGreen}{rgb}{0, 0.69, 0.31}
\definecolor{NavyBlue}{rgb}{0, 0.44, 0.75}
\newcommand{\hgreen}[1]{\textcolor{ForestGreen}{\textbf{#1}}} % highlight color
\usepackage{pifont}

\newcommand{\methodname}{Visual-RFT\xspace}
\definecolor{Cerulean}{rgb}{0.0, 0.48, 0.65}

\newcommand\blfootnote[1]{%
  \begingroup
  \renewcommand\thefootnote{}\footnote{#1}%
  \addtocounter{footnote}{-1}%
  \endgroup
}

\usepackage{amsmath}
\usepackage{CJKutf8}

\definecolor{iccvblue}{rgb}{0.21,0.49,0.74}
\usepackage[pagebackref,breaklinks,colorlinks,allcolors=iccvblue]{hyperref}

\def\paperID{*****} % *** Enter the Paper ID here
\def\confName{ICCV}
\def\confYear{2025}

\title{Visual-RFT: Visual Reinforcement Fine-Tuning}

\author{
Ziyu Liu$^{1,2}$\footnotemark[1] \quad
Zeyi Sun$^{1,2}$\footnotemark[1] \quad
Yuhang Zang$^{2}$\textsuperscript{\Letter}  \quad
Xiaoyi Dong$^{2,3}$ \quad
Yuhang Cao$^{2}$  \quad \\
Haodong Duan$^{2}$ \quad
Dahua Lin$^{2,3}$  \quad
Jiaqi Wang$^{2}$\textsuperscript{\Letter} \\
$^{1}$Shanghai Jiaotong University \quad $^{2}$Shanghai Artificial Intelligence Laboratory \\
$^3$The Chinese University of Hong Kong \quad  \\
{\tt\small \{liuziyu77, szy2023\}@sjtu.edu.cn, \{zangyuhang, wangjiaqi\}@pjlab.org.cn}\\
{\tt\small \url{https://github.com/Liuziyu77/Visual-RFT}}  
\vspace{-2mm}
}

\begin{document}
\begin{CJK}{UTF8}{gbsn}

\maketitle
\blfootnote{$^*$ Equal contribution.\ \ \textsuperscript{\Letter} Corresponding author.}

\begin{abstract}
Reinforcement Fine-Tuning (RFT) in Large Reasoning Models like OpenAI o1 learns from feedback on its answers, which is especially useful in applications when fine-tuning data is scarce.
Recent open-source work like DeepSeek-R1 demonstrates that reinforcement learning with verifiable reward is one key direction in reproducing o1.
While the R1-style model has demonstrated success in language models, its application in multi-modal domains remains under-explored.
This work introduces Visual Reinforcement Fine-Tuning (Visual-RFT), which further extends the application areas of RFT on visual tasks.
Specifically, Visual-RFT first uses Large Vision-Language Models (LVLMs) to generate multiple responses containing reasoning tokens and final answers for each input, and then uses our proposed visual perception verifiable reward functions to update the model via the policy optimization algorithm such as Group Relative Policy Optimization (GRPO).
We design different verifiable reward functions for different perception tasks, such as the Intersection over Union (IoU) reward for object detection.
Experimental results on fine-grained image classification, few-shot object detection, reasoning grounding, as well as open-vocabulary object detection benchmarks show the competitive performance and advanced generalization ability of Visual-RFT compared with Supervised Fine-tuning (SFT).
For example, \methodname improves accuracy by $24.3\%$ over the baseline in one-shot fine-grained image classification with around 100 samples.
In few-shot object detection, \methodname also exceeds the baseline by $21.9$ on COCO's two-shot setting and $15.4$ on LVIS.
Our \methodname represents a paradigm shift in fine-tuning LVLMs, offering a data-efficient, reward-driven approach that enhances reasoning and adaptability for domain-specific tasks.
\end{abstract}

\section{Introduction}

\begin{figure}[t]
    \begin{center}
    %\framebox[4.0in]{$\;$}
    \vspace{-12pt}
    \includegraphics[width=1.0\linewidth]{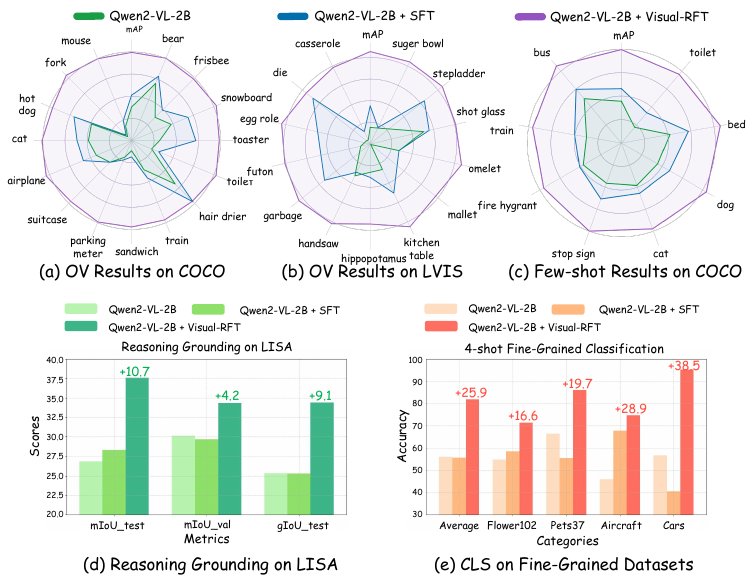}
    \end{center}
    \vspace{-12pt}
    \caption{\small Our \textbf{Visual} \textbf{R}einforcement \textbf{F}ine-\textbf{T}uning (\methodname) performs better than previous Supervised Fine-Tuning (SFT) on a variety of tasks, such as Open Vocabulary(OV)/Few-shot Detection, Reasoning Grounding, and Fine-grained Classification.}
    \label{fig:teaser}
    \vspace{-12pt}
\end{figure}

\begin{figure*}[t]
    \begin{center}
    %\framebox[4.0in]{$\;$}
    \vspace{-12pt}
    \includegraphics[width=1.\linewidth]{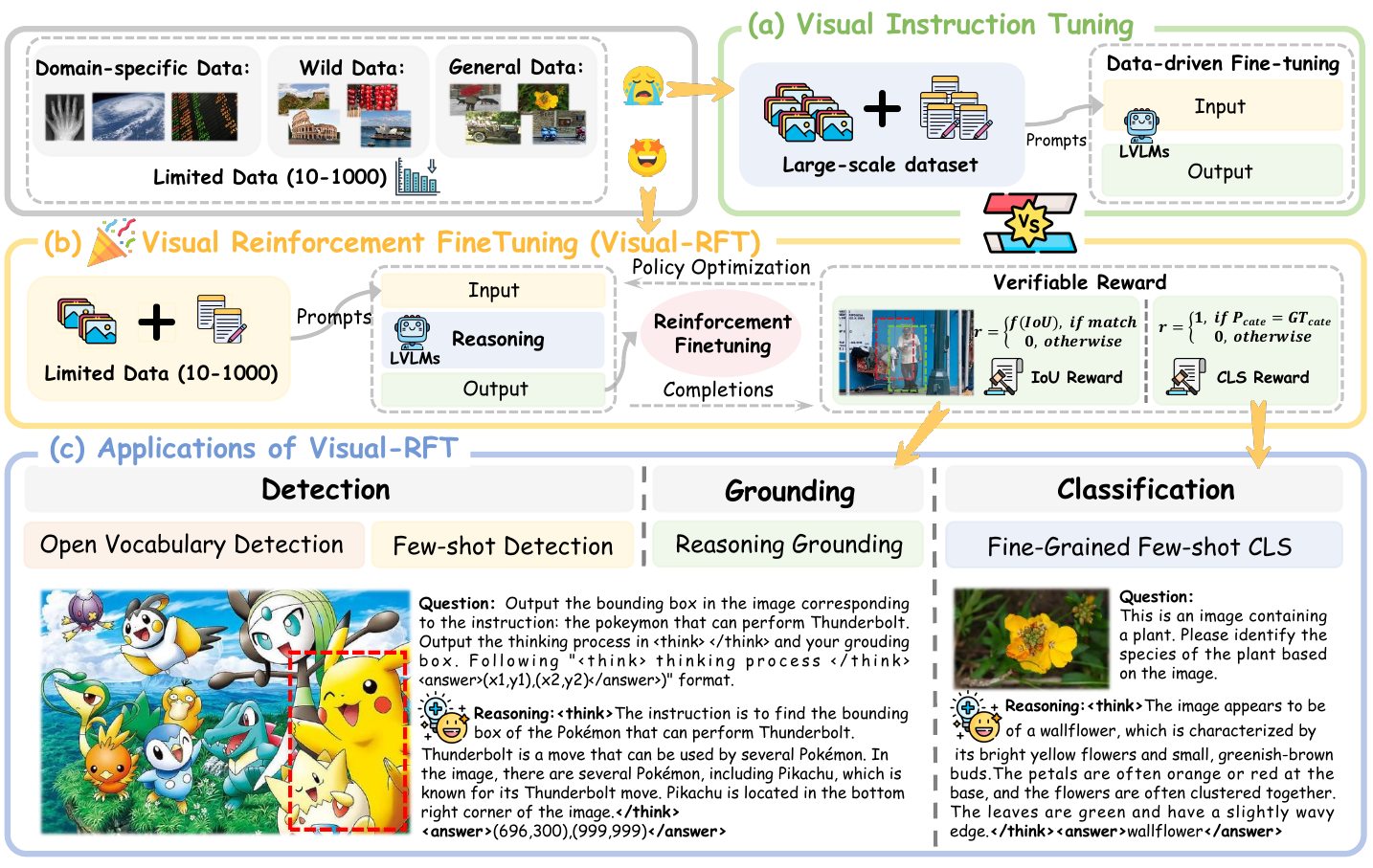}
    \end{center}
    \vspace{-12pt}
    \caption{\small \textbf{Overview of \methodname.}  Compared to the \textbf{(a)} Visual Instruction Tuning that is data-hungry, \textbf{(b)} our Visual Reinforcement Fine-Tuning (\methodname) is more data efficient with limited data.
    \textbf{(c)} We successfully empower Large Vision-Language Models (LVLMs) with RFT on a series of multi-modal tasks, and present examples of the model's reasoning process at the bottom.}
    \label{fig:teaser_framework}
    \vspace{-12pt}
\end{figure*}

% Building upon the success of Large Language Models \cite{2024gpt4o,dubey2024llama,wang2024qwen2}
Large Reasoning Models (LRMs) such as OpenAI o1 \cite{OpenAI_O1} represent frontier AI models designed to spend more time ``thinking'' before answering, and achieving excellent reasoning abilities.
One impressive capability of OpenAI o1 is \textbf{Reinforcement Fine-Tuning (RFT)} \footnote{\href{https://openai.com/form/rft-research-program}{https://openai.com/form/rft-research-program}}, which efficiently fine-tune the model with merely dozens to thousands of samples to excel at domain-specific tasks.
Although the implementation details of o1 are not publicly available, recent open-source studies like DeepSeek R1 \cite{DeepSeek-R1} reveal one promising direction in reproducing o1 is \textbf{Verifiable Rewards} \cite{lambert2024t,DeepSeek-R1,team2025kimi}: the reward score in reinforcement learning is directly determined by pre-defined rules, rather than predicted by the separate reward model \cite{ouyang2022training,liu2024skywork,zang2025internlm} trained on preference data.

A primary distinction between the RFT and Previous Supervised Fine-Tuning (SFT) lies in data efficiency.
Previous SFT paradigm (see \cref{fig:teaser_framework} (a)) directly imitates the ``ground-truth'' answers provided in the high-quality, curated data, thus relying on a large amount of training data.
By contrast, RFT evaluates the model's responses and adjusts based on whether they’re correct, helping it learn through trial and error.
Thus, RFT is particularly useful in domains where data is scarce \cite{OpenAI_O1,OpenAI_O3}.
However, a previous common sense is that RFT is applied merely in tasks like scientific (e.g., mathematics) and code generation.
That's because math and coding exhibit clear and objective final answers or test cases, making their rewards relatively straightforward to verify.
In this paper, we demonstrate that RFT can be applied \textbf{beyond} math and code domains to visual perception tasks.
Specifically, we introduce \textbf{Vi}sual \textbf{R}einforcement \textbf{F}ine-\textbf{T}uning (\textbf{\methodname}), which successfully extends RFT to empower Large Vision-Language Models (LVLMs) in various multi-modal tasks (see \cref{fig:teaser}), such as few-shot classification and open-vocabulary object detection.

To extend RFT on visual tasks, we present the implementation details of \methodname in \cref{fig:teaser_framework} (b).
For each input, \methodname uses Large Vision-Language Models (LVLMs) to generate multiple responses (trajectories) that contain the reasoning tokens and final answers.
Crucially, we define task-specific, rule-based verifiable reward functions to guide policy optimization, such as GRPO \cite{grpo}, in updating the model.
For instance, we propose the Intersection over Union (IoU) reward for the object detection task.
Our Visual-RFT contrasts with SFT, which relies on memorizing correct answers.
Instead, our approach explores different possible solutions and learns to optimize for a desired outcome defined by our verified reward function.
It's about discovering what works best, not just mimicking pre-defined answers.
Our approach shifts the training paradigm from data scaling in SFT to the strategic design of variable reward functions tailored to specific multi-modal tasks.
As shown in \cref{fig:teaser_framework} (c), the synergistic combination of verifiable rewards and \textbf{visual perception abilities} (e.g., detection, grounding, classification) allows our model to achieve rapid and data-efficient mastery of new concepts, facilitated by a detailed reasoning process.

\begin{figure*}[t]
    \begin{center}
    %\framebox[4.0in]{$\;$}
    \vspace{-12pt}
    \includegraphics[width=1.\linewidth]{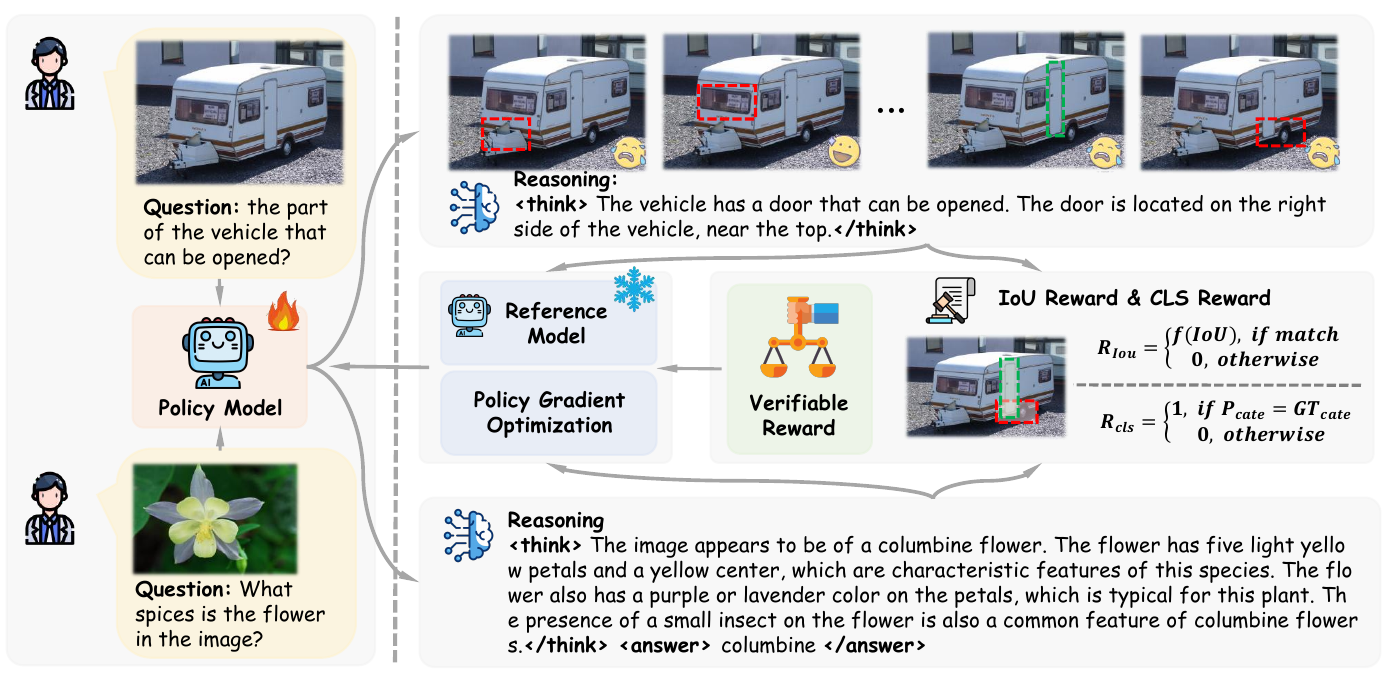}
    \end{center}
    \vspace{-12pt}
    \caption{\small \textbf{Framework of \methodname.} Given the question and visual image inputs, the policy model generates multiple responses containing reasoning steps. Then the verifiable reward such as IoU reward and CLS reward is used with the policy gradient optimization algorithm to update the policy model.}
    \label{fig:framework}
    \vspace{-12pt}
\end{figure*}

We validate the effectiveness of \methodname on the following tasks. In fine-grained image classification, the model utilizes its advanced reasoning capabilities to analyze fine-grained categories with high precision. In the one-shot setting with extremely limited data (e.g., around 100 samples), \methodname boosts the accuracy with $24.3\%$ over the baseline, while SFT dropped by $4.3\%$.
In few-shot experiments, \methodname also demonstrates exceptional performance with minimal training data, showcasing superior few-shot learning capabilities compared to SFT.
In reasoning grounding, \methodname excels in the LISA~\cite{lai2024lisa} dataset, which heavily relies on reasoning, outperforming specialized models like GroundedSAM~\cite{liu2024grounding}.
Furthermore, in open vocabulary object detection, \methodname quickly transfers recognition capabilities to new categories, including rare categories in LVIS~\cite{gupta2019lvis}, showing strong generalization.
Specifically, the 2B model achieves mAP improvements from $9.8$ to $31.3$ on new classes of COCO~\cite{coco} and from $2.7$ to $20.7$ on selected rare classes of LVIS~\cite{gupta2019lvis}.
These diverse visual perception tasks not only highlight \methodname's robust generalization capabilities in visual recognition but also underscore the crucial role of reinforcement learning in enhancing visual perception and reasoning.

In summary, our key contributions are as follows:

\noindent \textbf{(1)} We introduce Visual Reinforcement Fine-tuning (\methodname), which extends reinforcement learning with verifiable rewards on visual perception tasks that are effective with limited data for fine-tuning.

\noindent \textbf{(2)} We design different verifiable rewards for different visual tasks that enable efficient, high-quality reward computation at a negligible cost. This allows the seamless transfer of DeepSeek R1's style reinforcement learning to LVLMs.

\noindent \textbf{(3)} We conduct extensive experiments on various visual perception tasks, including fine-grained image classification, few-shot object detection, reasoning grounding, and open vocabulary object detection.
On all the settings, \methodname achieves remarkable performance improvements, significantly surpassing the supervised fine-tuning baselines.

\noindent \textbf{(4)} We fully \textit{open-source} the training code, training data, and evaluation scripts on \href{https://github.com/Liuziyu77/Visual-RFT}{Github} to facilitate further research.

\label{sec_1_introduction}

\section{Related Work}
\vspace{2mm}
\noindent \textbf{Large Vision Language Models} (LVLMs) like GPT-4o~\cite{2024gpt4o} achieves excellent visual understanding ability by integrating both visual and textual data. This integration enhances the models' ability to understand complex multi-modal inputs and enables more advanced AI systems~\cite{wang2024qwen2,li2024llavaov,zhang2024internlm,liu2024deepseekv3} capable of processing and responding to both images and text. Generally, the training of LVLMs involves two steps: (a) pre-training and (b) post-training which contains supervised fine-tuning and reinforcement learning. Post-training is crucial in improving the model's response quality, instruction following, and reasoning abilities. While there has been significant research on using reinforcement learning to enhance LLMs during post-training~\cite{lm-human-preferences, Learning-to-summarize, Training-language-models, RL4LMs, zang2024contextual, Grounding-large-language-models, Aligning-LLMs-with-RLHF, ILQL, LMRL-Gym, ArCHer, ReAct}, the progress for LVLMs has been slower. In this paper, we propose \methodname, which used GRPO-based reinforcement algorithms and verifiable reward during the post-training phase to enhance the model’s visual perception and reasoning capabilities.

\noindent \textbf{Reinforcement Learning} Recently, with the emergence of reasoning models like OpenAI's o1~\cite{OpenAI_O1}, the research focus in Large Language Models (LLMs) has increasingly shifted towards enhancing the models' reasoning capabilities through reinforcement learning (RL) techniques. Studies have explored improving LLMs' performance in reasoning tasks such as solving mathematical problems~\cite{grpo,yang2024qwen2math,ying2024internlmmath,cai2024internlm2,luong2024reft} and coding~\cite{hui2024qwen2coder,jiao2024preferencecode,zhang2024o1,zhang2024codedpo}. A notable breakthrough in this area is Deepseek-R1-Zero~\cite{DeepSeek-R1}, which introduced a new approach to achieving robust reasoning capabilities using RL merely, eliminating the supervised fine-tuning (SFT) stage.
However, current research on RL-based reasoning has largely been confined to the language domain, with limited exploration of its application in multi-modal settings. For LVLMs, RL has primarily been used for tasks like mitigating hallucinations and aligning models with human preference~\cite{llavarlhf,hadpo,povid,sun2023aligning,yu2024rlhfv,liu2024mia,yu2024rlaif,zhou2024aligning}, but there remains a significant gap in research focusing on enhancing reasoning and visual perception of Large Vision Language Models. To address this gap, our work introduces a novel reinforcement fine-tuning strategy \methodname, applying verifiable rewards with GRPO-based~\cite{grpo} RL to a broad range of visual perception tasks. Our approach aims to improve the performance of LVLMs in processing various visual tasks, especially when the fine-tuning data is limited.
\label{sec_2_related_work}

\section{Methodology}
\subsection{Preliminary}

\begin{figure*}[t]
    \begin{center}
    \includegraphics[width=.96\linewidth]{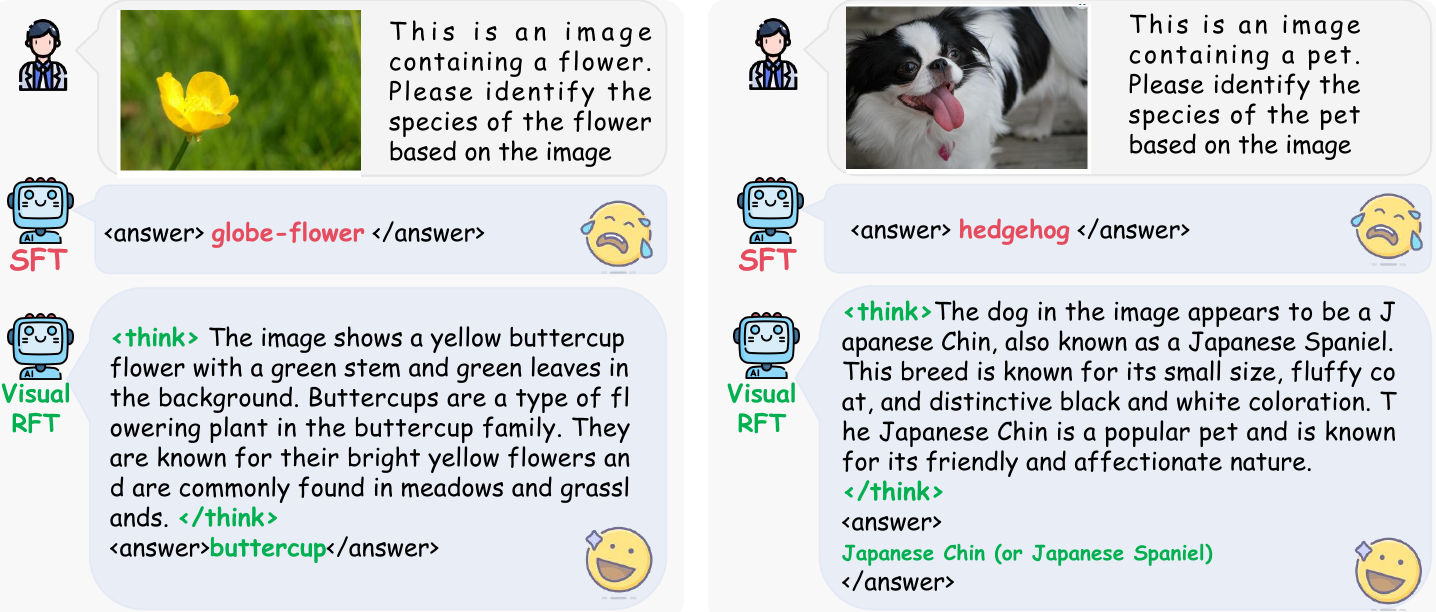}
    \end{center}
    \vspace{-6mm}
    \caption{\small \textbf{Qualitative results of Fine-Grained Image Classification.} The thinking process significantly improves the reasoning ability of LVLMs, leading to higher image classification performance.}
    \label{fig:cls_case}

\end{figure*}

\begin{figure*}[t]
    \begin{center}
    %\framebox[4.0in]{$\;$}
    \includegraphics[width=.96\linewidth]{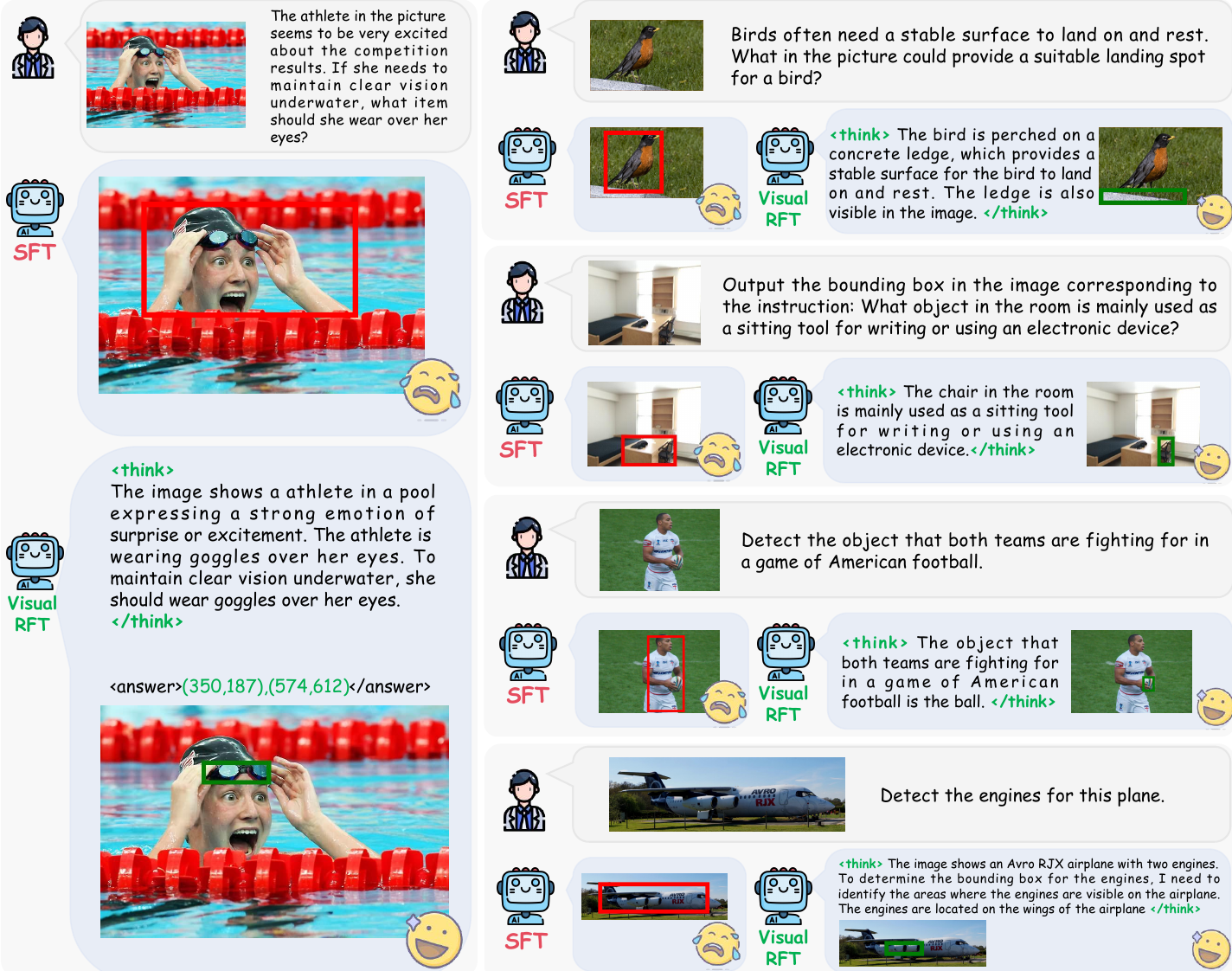}
    \end{center}
    \caption{\small \textbf{Qualitative results of reasoning grounding on LISA~\cite{lai2024lisa}.} Thinking process significantly improves reasoning grounding ability with \methodname.}
    \label{fig:lisa_case}
\end{figure*}

\begin{table}[t]
\caption{
\small
\textbf{Prompts used to construct the dataset.} We have listed the detection prompt and classification prompt separately.}
\vspace{-6mm}
\label{tab:prompt}
\begin{center}
\setlength{\tabcolsep}{4pt}
\scalebox{0.9}
{
\begin{tabular}{p{9cm}}  % 设置列宽，单位为cm，可以根据需要调整
\toprule
\textbf{Detection Prompt:} Detect all objects belonging to the category '$\{$category$\}$' in the image, and provide the bounding boxes (between 0 and 1000, integer) and confidence (between 0 and 1, with two decimal places). If no object belonging to the category '$\{$category$\}$' in the image, return 'No Objects'. Output the thinking process in $<$think$>$ $<$/think$>$ and final answer in $<$answer$>$ $<$/answer$>$ tags. The output answer format should be as follows: $<$think$>$ ... $<$/think$>$$<$answer$>$[{'Position': [x1, y1, x2, y2], 'Confidence': number}, ...]$<$/answer$>$ Please strictly follow the format.
\\
\midrule
\textbf{Classification Prompt:} This is an image containing a plant. Please identify the species of the plant based on the image. Output the thinking process in $<$think$>$ $<$/think$>$ and final answer in $<$/think$>$ $<$/answer$>$ tags. The output answer format should be as follows: $<$think$>$ ... $<$/think$>$ $<$/think$>$species name$<$/answer$>$ Please strictly follow the format.
\\
\bottomrule
\end{tabular}
}
\vspace{-4mm}
\end{center}
\end{table}

\paragraph{Reinforcement Learning with Verifiable Rewards.}
Reinforcement Learning with Verifiable Rewards (RLVR) \cite{lambert2024t,DeepSeek-R1,team2025kimi} is a novel training approach designed to enhance language models in tasks with objectively verifiable outcomes, such as math and coding. Unlike previous Reinforcement Learning from Human Feedback (RLHF) \cite{ouyang2022training,liu2024skywork,zang2025internlm}, which relies on a trained reward model, RLVR instead uses a direct verification function to assess correctness.
This approach simplifies the reward mechanism while maintaining strong alignment with the task's inherent correctness criteria.
% By using verifiable answer matching, RLVR refines model reasoning capabilities across multiple domains. It builds upon existing techniques that bootstrap reasoning in language models, offering a streamlined alternative to reinforcement learning with verifiable feedback. 
Given the input question $q$, the policy model $\pi_{\theta}$ generates responses $o$ and receives the verifiable reward. More specifically, RLVR optimizes the following objective:
\begin{align}
    & \max_{\pi_{\theta}} \mathbb{E}_{o \sim \pi_{\theta}(q)} \left[ R_{\text{RLVR}}(q, o) \right] \\
    & = \left[ R(q, o) - \beta \text{KL}[\pi_{\theta}(o|q) \parallel \pi_{\text{ref}}(o|q)] \right],
\end{align}
 where $\pi_{\text{ref}}$ is the reference model before optimization, $R$ is the verifiable reward function, and $\beta$ is the hyper-parameters to control the KL-divergence. The verifiable reward function $R$ takes the question and output pair $(q,o)$ as inputs, and checks if the ground-truth answer remains the same as the prediction $o$:
\begin{equation}
    R(q, o) = 
    \begin{cases} 
    1 ,& \text{if o = ground truth}, \\
    0 ,& \text{otherwise}.
    \end{cases} 
\end{equation}
\paragraph{DeepSeek R1-Zero and GRPO.}
The DeepSeek R1-Zero algorithm eliminates dependence on supervised fine-tuning (SFT) by employing reinforcement learning for training, specifically through its Group Relative Policy Optimization (GRPO) framework. Different from reinforcement learning algorithms such as PPO \cite{PPO} that require a critic model to evaluate policy performance, GRPO compares groups of candidate responses directly, eliminating the need for an additional critic model. For a given question $q$, GRPO first generates $G$ distinct responses $\{o_1, o_2, ..., o_G\}$ from the current policy $\pi_{\theta_{\text{old}}}$.
Then GRPO takes actions based on these responses and denotes the obtained rewards as $\{r_1, r_2, ..., r_G\}$. By computing their mean and standard deviation for normalization, GRPO determines the relative quality of these responses:
\begin{align}
A_i &= \frac{r_i - \text{mean}(\{r_1, \dots, r_G\})}{\text{std}(\{r_1, \dots, r_G\})},
\end{align}
where $A_i$ represents the relative quality of the $i$-th answer. 
GRPO encourages the model to favor better answers with a high reward value within the group.
% , represented by the log ratios of predicted probability $\pi_{\theta}(o_i \mid q) / \pi_{\theta_{\text{old}}}(o_i \mid q)$.
% However, we do not amplify this ratio uncontrollably; instead, we apply a clipping operation to constrain its value within $1 - \epsilon$ and $1 + \epsilon$, where $\epsilon$ is the clip ratio. 

% Therefore, the final optimization process is:

% \begin{align}
% J_{\text{GRPO}}(\theta) &= \mathbb{E} \bigg[ \sum_{i=1}^{G} \min \bigg(  
% \frac{\pi_{\theta}(o_i \mid q)}{\pi_{\theta_{\text{old}}}(o_i \mid q)} A_i,  \notag \\
% &\quad \text{clip} \left( \frac{\pi_{\theta}(o_i \mid q)}{\pi_{\theta_{\text{old}}}(o_i \mid q)}, 1 - \epsilon, 1 + \epsilon \right) A_i \bigg) \notag \\
% &\quad - \beta D_{\text{KL}}(\pi_{\theta} \parallel \pi_{\text{ref}}) \bigg]
% \end{align}

% \begin{align}
% D_{\text{KL}} &= \pi_{\text{ref}}(o_i \mid q) \left( \log \frac{\pi_{\text{ref}}(o_i \mid q)}{\pi_{\theta}(o_i \mid q)} - 1 \right)
% \end{align}

\subsection{\methodname}
The framework of \methodname is shown in Fig.~\ref{fig:framework}. The multi-modal input data from the user consists of images and questions. The policy model $\pi_{\theta}$ outputs a reasoning process and generates a group of responses based on the input. Each response is passed through a verifiable reward function to compute the reward. After group computation of the rewards for each output, the quality of each response is evaluated and used to update the policy model. To ensure the stability of the policy model training, \methodname uses KL divergence to limit the difference between the policy model and the reference model.
We will further discuss how to design the verifiable reward for visual tasks in \cref{sec:reward}, and the data preparation steps in \cref{sec:data_prepare}

\subsubsection{Verifiable Reward in Visual Perception}\label{sec:reward}
The reward model is a key step in reinforcement learning (RL) that aligns models with preference alignment algorithms, which can be as straightforward as a verification function that checks for exact matches between predictions and ground-truth answers.
The RL training process in the recent DeepSeek-R1~\cite{DeepSeek-R1} model achieves a significant improvement in the model's reasoning ability through the verifiable reward design.
To transfer this strategy to the visual domain, we design different rule-based verifiable reward functions for various visual perception tasks.

\paragraph{IoU Reward in Detection Tasks.}
For the detection task, the model's output consists of bounding boxes (bbox) and corresponding confidences.
The reward function for the detection task should adequately consider the Intersection-over-Union (IoU) metric, which is used to compute the mean Average Precision (mAP) in evaluation.
Therefore, we design an IoU and confidence-based reward function $R_d$. First, for the model's output box and confidence, we sort these boxes based on their confidence, denoted as $\{b_1, b_2, ..., b_n\}$. We then match each $b_i$ with the ground truth bbox,$\{b^g_1, b^g_2, ..., b^g_m\}$, and calculate the IoU, while setting an IoU threshold $\tau$. Bounding boxes with an IoU below this threshold $\tau$ are considered invalid, and unmatched bboxes have an IoU of 0. After matching, we obtain the IoU and confidence for each box from the initial set, denoted as $\{iou_1: c_1, iou_2: c_2, ..., iou_n: c_n\}$.

We then use these IoU results and confidence to construct our reward $R_d$. Our reward $R_d$ consists of three parts, including the IoU reward, Confidence reward, and Format reward:
\begin{align}
R_{d} = R_{\text{IoU}} + R_{\text{conf}} + R_{\text{format}}.
\end{align}
The IoU reward $R_{\text{IoU}}$ is the average IoU of all the bounding boxes in the model's output,
\begin{align}
R_{\text{IoU}} = \frac{iou_1+iou_2+...+iou_n}{n}.
\end{align}
The confidence reward $R_{\text{conf}}$ is related to IoU. For each bounding box, if the $iou_i$ is non-zero, indicating a successful match, the confidence reward for this single box $r_c$ as the predicted confidence in computed as:
\begin{align}
r_{ci} = \left\{
\begin{array}{ll}
c_i & , \text{if} \quad iou_i \neq 0,\\
1-c_i & , \text{if} \quad iou_i = 0. \\
\end{array}
\right.
\end{align}
This means that for successfully matched boxes, the higher the confidence, the better. If the $iou_i$ is zero, indicating a failed match, the lower the confidence reward $r_c$ for this box, the better. The overall confidence reward $R_{\text{conf}}$ for the model's output is the average of the confidence rewards of all the bounding boxes in that output,
\begin{align}
R_{\text{conf}} = \frac{ \sum_{i=1}^{n} r_{ci}}{n}.
\end{align}
The format reward $R_{\text{format}}$ is used to force the model prediction to meet the required HTML tag format of $<$think$>$ and $<$answer$>$ (will detailed in \cref{sec:data_prepare}).

\paragraph{CLS Reward in Classification Tasks.}
In classification tasks, the reward function we use consists of two parts: accuracy reward $R_{\text{acc}}$ and format reward $R_{\text{format}}$.
The accuracy reward is determined by comparing the model's output class with the ground truth class, yielding a value of 1 for correct classification and 0 for incorrect classification:
\begin{align}
R_{\text{cls}} = R_{\text{acc}} + R_{\text{format}}.
\end{align}

\subsubsection{Data Preparation}\label{sec:data_prepare}
To train the \methodname on various visual perception tasks, we need to construct the multi-modal training dataset.
Similar to DeepSeek-R1, to enhance the model's reasoning ability and apply this ability to improve visual perception, \methodname designed a prompt format to guide the model to output its reasoning process before providing the final answer.
The prompts used for detection and classification tasks are shown in Tab~\ref{tab:prompt}.

During the training process, we use the format reward to guide the model to output its reasoning process and the final answer in a structured format.
The reasoning process is key to the model's self-learning and improvement during reinforcement fine-tuning, while the reward determined by the answer directs the model's optimization.
\label{sec_3_methodology}

\section{Experiments}
\subsection{Experimental Setup}

\begin{table}[t]
\caption{
\small
\textbf{Few-shot results on Fine-grained Classification dataset.} We evaluated four fine-grained image classification datasets.}

\vspace{-6mm}
\label{tab:cls}
\begin{center}
\setlength{\tabcolsep}{4pt}
\scalebox{0.96}
{
\begin{tabular}{l|l|cccc}
\toprule
\multicolumn{1}{c}{\bf Models}&\multicolumn{1}{c}{\bf Average} &\rotatebox{90}{ Flower102} &\rotatebox{90}{ Pets37} &\rotatebox{90}{ FGVC} &\rotatebox{90}{Cars196}
\\
% \midrule
% \rowcolor[HTML]{F2F2F2} \multicolumn{8}{c}{\textbf{\textit{Qwen2-VL-2B}}} \\
\midrule
Qwen2-VL-2B  & 56.0 & 54.8 & 66.4 & 45.9 & 56.8 \\

\midrule
\multicolumn{6}{c}{\textbf{\textit{one-shot}}} \\
\midrule
$+$ SFT & 51.7 & 56.6 & 54.7 & 65.3 & 30.0 \\
\rowcolor[HTML]{DAEFF9} $+$ \methodname & 80.3 & 70.8 & 84.1 & 72.5 & 93.8 \\
$\Delta$ & \hgreen{+24.3} & \hgreen{+16.0} & \hgreen{+17.7} & \hgreen{+26.6} & \hgreen{+37.0} \\

\midrule
\multicolumn{6}{c}{\textbf{\textit{2-shot}}} \\
\midrule
$+$ SFT & 58.8 & 60.3 & 65.6 & 68.9 & 40.2 \\
\rowcolor[HTML]{DAEFF9} $+$ \methodname & 83.5 & 75.8 & 87.5 & 75.3 & 95.4 \\
$\Delta$ & \hgreen{+27.5} & \hgreen{+21.0} & \hgreen{+21.1} & \hgreen{+29.4} & \hgreen{+38.6} \\

\midrule
\multicolumn{6}{c}{\textbf{\textit{4-shot}}} \\
\midrule
$+$ SFT & 55.6 & 58.5 & 55.5 & 67.9 & 40.5 \\
\rowcolor[HTML]{DAEFF9} $+$ \methodname & 81.9 & 71.4 & 86.1 & 74.8 & 95.3 \\
$\Delta$ & \hgreen{+25.9} & \hgreen{+16.6} & \hgreen{+19.7} & \hgreen{+28.9} & \hgreen{+38.5} \\

\midrule
\multicolumn{6}{c}{\textbf{\textit{8-shot}}} \\
\midrule
$+$ SFT & 60.3 & 59.6 & 71.4 & 69.2 & 40.9 \\
\rowcolor[HTML]{DAEFF9} $+$ \methodname & 85.1 & 77.7 & 90.2 & 75.9 & 96.5 \\
$\Delta$ & \hgreen{+29.1} & \hgreen{+22.9} & \hgreen{+23.8} & \hgreen{+30.0} & \hgreen{+39.7} \\

\midrule
\multicolumn{6}{c}{\textbf{\textit{16-shot}}} \\
\midrule
$+$ SFT & 64.0 & 66.8 & 71.6 & 76.1 & 41.5 \\
\rowcolor[HTML]{DAEFF9} $+$ \methodname & 85.3 & 79.2 & 87.1 & 79.4 &  95.3\\
$\Delta$ & \hgreen{+29.3} & \hgreen{+24.4} & \hgreen{+20.7} & \hgreen{+33.5} &  \hgreen{+38.5}\\
\bottomrule
\end{tabular}
}
\vspace{-4mm}
\end{center}
\end{table}

\begin{table}[htbp]
\caption{
\small
\textbf{Few-Shot results on COCO dataset of 8 categories.} We conducted one-shot, 2-shot, 4-shot, 8-shot, and 16-shot experiments on 8 categories from the COCO dataset.}
\vspace{-6mm}
\label{tab:few-shot_coco}
\begin{center}
\setlength{\tabcolsep}{2pt}
\scalebox{0.77}
{
\begin{tabular}{l|l|cccccccc}
\toprule
\multicolumn{1}{c}{\bf Models} &\multicolumn{1}{c}{\bf mAP} &\rotatebox{90}{\bf bus} &\rotatebox{90}{\bf train} &\rotatebox{90}{\bf fire hydrant} &\rotatebox{90}{\bf stop sign} &\rotatebox{90}{\bf cat}&\rotatebox{90}{\bf dog}&\rotatebox{90}{\bf bed}&\rotatebox{90}{\bf toilet}
\\
\midrule
\rowcolor[HTML]{F2F2F2} \multicolumn{10}{c}{\textbf{\textit{Qwen2-VL-2B}}} \\
\midrule
Baseline & 19.6 & 19.0 & 15.8 & 25.8 & 18.4 & 29.9 & 23.2 & 14.6 & 9.8 \\
\midrule
\multicolumn{10}{c}{\textbf{\textit{1-shot}}} \\
\midrule
$+$ SFT  & 19.5 & 18.3 & 17.4 & 23.1 & 18.2 & 28.0 & 23.4 & 17.3 & 10.4 \\
\rowcolor[HTML]{DAEFF9} $+$ \methodname & 33.6 & 23.4 & 35.7 & 39.1 & 23.8 & 54.3 & 42.5 & 19.5 & 30.8 \\
$\Delta$ & \hgreen{+14.0} & \hgreen{+4.4} & \hgreen{+19.9} & \hgreen{+13.3} & \hgreen{+5.4} & \hgreen{+24.4} & \hgreen{+19.3} & \hgreen{+4.9} & \hgreen{+21.0}\\
\midrule
\multicolumn{10}{c}{\textbf{\textit{2-shot}}} \\
\midrule
$+$ SFT  & 21.0 & 22.1 & 15.8 & 29.8 & 19.0 & 28.9 & 26.5 & 15.5 & 10.2 \\
\rowcolor[HTML]{DAEFF9} $+$ \methodname & 41.5 & 28.8 & 39.6 & 38.2 & 48.0 & 63.8 & 52.7 & 25.9 & 34.9 \\
$\Delta$ & \hgreen{+21.9} & \hgreen{+9.8} & \hgreen{+23.8} & \hgreen{+12.4} & \hgreen{+29.6} & \hgreen{+33.9} & \hgreen{+29.5} & \hgreen{+11.3} & \hgreen{+25.1}\\
\midrule
\multicolumn{10}{c}{\textbf{\textit{4-shot}}} \\
\midrule
$+$ SFT  & 25.2 & 25.4 & 23.6 & 26.6 & 21.5 & 35.6 & 30.6 & 18.4 & 19.9 \\
\rowcolor[HTML]{DAEFF9} $+$ \methodname & 40.6 & 30.0 & 40.6 & 45.7 & 35.0 & 60.9 & 44.9 & 24.6 & 43.1 \\
$\Delta$ & \hgreen{+21.0} & \hgreen{+11.0} & \hgreen{+24.8} & \hgreen{+19.9} & \hgreen{+16.6} & \hgreen{+31.0} & \hgreen{+21.7} & \hgreen{+10.0} & \hgreen{+33.3}\\
\midrule
\multicolumn{10}{c}{\textbf{\textit{8-shot}}} \\
\midrule
$+$ SFT & 30.2 & 25.8 & 35.1 & 29.4 & 21.9 & 44.5 & 39.0 & 22.6 & 23.5 \\
\rowcolor[HTML]{DAEFF9} $+$ \methodname & 47.4 & 36.2 & 47.9 & 50.4 & 47.7 & 65.2 & 57.0 & 30.4 & 44.0 \\
$\Delta$ & \hgreen{+27.8} & \hgreen{+17.2} & \hgreen{+32.1} & \hgreen{+24.6} & \hgreen{+29.3} & \hgreen{+35.3}& \hgreen{+33.8} & \hgreen{+15.8} & \hgreen{+34.2}\\
\midrule
\multicolumn{10}{c}{\textbf{\textit{16-shot}}} \\
\midrule
$+$ SFT & 31.3 & 24.0 & 35.9 & 32.0 & 23.6 & 39.8 & 40.6 & 27.5 & 26.8 \\
\rowcolor[HTML]{DAEFF9} $+$ \methodname & 46.8 & 36.2 & 44.4 & 51.4 & 48.5 & 66.6 & 56.2 & 27.6 & 43.4 \\
$\Delta$ & \hgreen{+27.2} & \hgreen{+17.2} & \hgreen{+28.6} & \hgreen{+25.6} & \hgreen{+30.1} & \hgreen{+36.7}& \hgreen{+33.0} & \hgreen{+13.0} & \hgreen{+33.6}\\
\midrule
\rowcolor[HTML]{F2F2F2} \multicolumn{10}{c}{\textbf{\textit{Qwen2-VL-7B}}} \\
\midrule
Baseline & 43.0 & 35.0 & 43.3 & 37.1 & 36.7 & 57.3 & 50.3 & 37.4 & 47.1  \\
\midrule
\multicolumn{10}{c}{\textbf{\textit{4-shot}}} \\
\midrule
$+$ SFT & 44.1 & 41.4 & 51.7 & 35.6 & 30.8 & 60.5 & 52.7 & 38.5 & 41.5 \\
\rowcolor[HTML]{DAEFF9} $+$ \methodname & 54.3 & 44.3 & 59.8 & 52.0 & 46.0 & 72.7 & 62.8 & 41.9 & 55.0 \\
$\Delta$ & \hgreen{+11.3} & \hgreen{+9.3} & \hgreen{+16.5} & \hgreen{+14.9} & \hgreen{+9.3} & \hgreen{+15.4} & \hgreen{+12.5} & \hgreen{+4.5} & \hgreen{+7.9} \\

\bottomrule
\end{tabular}
}
\vspace{-4mm}
\end{center}
\end{table}

\begin{table}[htbp]
\caption{
\small
\textbf{Few-shot results on LVIS dataset of 6 rare categories.} We conducted 10-shot experiments on 6 rare categories from the LVIS dataset.}
\vspace{-6mm}
\label{tab:few-shot_lvis}
\begin{center}
\setlength{\tabcolsep}{2pt}
\scalebox{0.95}
{
\begin{tabular}{l|l|cccccc}
\toprule
\multicolumn{1}{c}{\bf Models} &\multicolumn{1}{c}{\bf mAP} &\rotatebox{90}{\bf horse buggy} &\rotatebox{90}{\bf die} &\rotatebox{90}{\bf kitchen table} &\rotatebox{90}{\bf omelet}&\rotatebox{90}{\bf papaya}&\rotatebox{90}{\bf stepladder}
\\
\midrule
% GroudingDINO-B~\cite{liu2024grounding} & 36.1 & 22.5 & 0.0 & 92.5 & 26.6 & 34.2 & 41.0\\
% \midrule
% \rowcolor[HTML]{F2F2F2} \multicolumn{8}{c}{\textbf{\textit{Qwen2-VL-2B}}} \\
% \midrule
Qwen2-VL-2B & 4.0 & 2.9 & 1.2 & 13.4 & 4.7 & 1.5 & 0.0\\
\midrule
% \multicolumn{8}{c}{\textbf{\textit{one-shot}}} \\
% \midrule
$+$ SFT & 10.0 & 7.0 & 7.6 & 34.1 & 4.7 & 6.3 & 0.0\\
\rowcolor[HTML]{DAEFF9}  $+$ \methodname & 19.4 & 9.1 & 19.6 & 42.2 & 20.4 & 14.5 & 10.9 \\
$\Delta$ & \hgreen{+15.4} & \hgreen{+6.2} & \hgreen{+18.4} & \hgreen{+29.2} & \hgreen{+15.7}& \hgreen{+13.0}& \hgreen{+10.9}\\
% \midrule
% \multicolumn{8}{c}{\textbf{\textit{2-shot}}} \\
% \midrule
% $+$ SFT & 16.5 & 7.1 & 9.5 & 63.2 & 12.9 & 6.4 & 0.0\\
% \rowcolor[HTML]{DAEFF9}  $+$ \methodname & 29.4 & 10.3 & 20.9 & 91.4 & 20.4 & 14.1 & 19.3\\
% $\Delta$ & \hgreen{+20.9} & \hgreen{+7.1} & \hgreen{+3.1} & \hgreen{+72.5} & \hgreen{+9.2}& \hgreen{+14.1}& \hgreen{+19.3}\\
% \midrule
% % \multicolumn{8}{c}{\textbf{\textit{4-shot}}} \\
% % \midrule
% $+$ SFT & 16.5 & 7.1 & 9.5 & 63.2 & 12.9 & 6.4 & 0.0\\
% \rowcolor[HTML]{DAEFF9}  $+$ \methodname & 29.4 & 10.3 & 20.9 & 91.4 & 20.4 & 14.1 & 19.3\\
% $\Delta$ & \hgreen{+20.9} & \hgreen{+7.1} & \hgreen{+3.1} & \hgreen{+72.5} & \hgreen{+9.2}& \hgreen{+14.1}& \hgreen{+19.3}\\
% \midrule
% \rowcolor[HTML]{F2F2F2} \multicolumn{8}{c}{\textbf{\textit{Qwen2-VL-7B}}} \\
\midrule
Qwen2-VL-7B & 15.4 & 19.7 & 21.9 & 14.5 & 18.1 & 18.5 & 0.0 \\
\midrule
% \multicolumn{8}{c}{\textbf{\textit{4-shot}}} \\
% \midrule
$+$ SFT & 27.6 & 26.9 & 21.9 & 49.7 & 29.2 & 25.2 & 12.7 \\
\rowcolor[HTML]{DAEFF9}  $+$ \methodname & 33.8 & 26.2 & 27.8 & 70.6 & 23.5 & 21.2 & 29.3 \\
$\Delta$ & \hgreen{+18.4} & \hgreen{+6.5} & \hgreen{+5.9} & \hgreen{+56.1} & \hgreen{+5.4}& \hgreen{+2.7}& \hgreen{+29.3}\\
\bottomrule
\end{tabular}
}
\vspace{-4mm}
\end{center}
\end{table}

\begin{table}[htbp]
\caption{
\small
\textbf{Few-shot results on MG dataset of 5 categories.} By introducing out-of-domain data, we increased the difficulty of model recognition and reasoning, further demonstrating the strong generalization ability of reinforcement fine-tuning in visual perception tasks.}
\vspace{-6mm}
\label{tab:mg_result}
\begin{center}
\setlength{\tabcolsep}{2pt}
\scalebox{0.80}
{
\begin{tabular}{l|l|ccccc}
\toprule
\multicolumn{1}{c}{\bf Models} &\multicolumn{1}{c}{\bf mAP} &\multicolumn{1}{c}{\bf bird} &\multicolumn{1}{c}{\bf feline-or-canid} &\multicolumn{1}{c}{\bf serpent} &\multicolumn{1}{c}{\bf slime} &\multicolumn{1}{c}{\bf wyvern}
\\
% \midrule
% \rowcolor[HTML]{F2F2F2} \multicolumn{8}{c}{\textbf{\textit{Qwen2-VL-2B}}} \\
\midrule
Qwen2-VL-2B & 20.6 & 12.9 & 19.8 & 25.5 & 18.4 & 26.4 \\
\midrule
\multicolumn{7}{c}{\textbf{\textit{4-shot}}} \\
\midrule
$+$ SFT & 26.8 & 19.5 & 22.4 & 26.8 & 33.5 & 31.8 \\
\rowcolor[HTML]{DAEFF9} $+$ \methodname & 61.8 & 63.9 & 53.2 & 70.2 & 64.5 & 57.5 \\
$\Delta$ & \hgreen{+41.2} & \hgreen{+51.0} & \hgreen{+33.4} & \hgreen{+44.7} & \hgreen{+46.1} & \hgreen{+31.1} \\
\midrule
\multicolumn{7}{c}{\textbf{\textit{16-shot}}} \\
\midrule
$+$ SFT & 51.3 & 42.7 & 44.4 & 56.4 & 61.1 & 52.0 \\
\rowcolor[HTML]{DAEFF9} $+$ \methodname & 63.4 & 59.9 & 50.8 & 76.3 & 71.7 & 58.1\\
$\Delta$ & \hgreen{+42.8} & \hgreen{+47.0} & \hgreen{+56.4} & \hgreen{+50.8} & \hgreen{+53.3} & \hgreen{+31.7} \\

% \midrule
% \rowcolor[HTML]{F2F2F2} \multicolumn{8}{c}{\textbf{\textit{Qwen2-VL-7B}}} \\
% \midrule
% Baseline  & w & 9.6 & 11.8 & 2.8 & 7.3 & 18.6 & 7.5\\
% \midrule
% \multicolumn{8}{c}{\textbf{\textit{4-shot}}} \\
% \midrule
% $+$ SFT & w/o  & 42.6 & 39.4 & 41.5 & 27.3 & 50.2 & 54.8  \\
% \rowcolor[HTML]{DAEFF9} $+$ RL & w  & 62.2 & 64.6 & 57.5 & 66.6 & 64.0 & 58.3\\
% $\Delta$ & - & \hgreen{+52.6} & \hgreen{+52.8} & \hgreen{+54.7} & \hgreen{+59.3} & \hgreen{+45.4}  & \hgreen{+50.8} \\

% \midrule
% \multicolumn{8}{c}{\textbf{\textit{16-shot}}} \\
% \midrule
% $+$ SFT & w/o  & 56.2 & 44.5 & 52.4 & 55.8 & 73.8 & 54.3 \\
% \rowcolor[HTML]{DAEFF9} $+$ RL & w  & 64.5 & 63.8 & 59.0 & 70.8 & 68.8 & 59.8 \\
% $\Delta$ & - & \hgreen{+54.9} & \hgreen{+52.0} & \hgreen{+56.2} & \hgreen{+63.5} & \hgreen{+50.2} & \hgreen{+52.3}  \\

\bottomrule
\end{tabular}
}
\end{center}
\end{table}

% Table generated by Excel2LaTeX from sheet 'Sheet1'
\begin{table}[htbp]
    \caption{\textbf{Reasoning Grounding Results on LISA~\cite{lai2024lisa}}. \methodname surpasses SFT in reasoning grounding with 239 training images.}
  \centering
    \begin{tabular}{lccc}
    \toprule
    Model  & mIoU$_{\text{test}}$ & mIoU$_{\text{val}}$ & gIoU$_{\text{test}}$\\
    \midrule
    OV-Seg~\cite{liang2023open} & 28.4 & 30.5 & 26.1 \\
    X-Decoder~\cite{zou2023generalized} & 28.5 & 29.1 &  24.3 \\
    GroundedSAM~\cite{liu2024grounding} & 26.2 & 28.6 & 21.3 \\
    \midrule
    Qwen2-VL-2B & 26.9 & 30.1 & 25.3 \\
    \midrule
    $+$ SFT   & 28.3 & 29.7 & 25.3 \\
    \rowcolor[HTML]{DAEFF9} $+$ \methodname   & 37.6 & 34.4 & 34.4 \\
    $\Delta$ & \hgreen{+10.7} & \hgreen{+4.3} & \hgreen{+9.1} \\
    \midrule
    Qwen2-VL-7B & 40.4 & 45.2 & 38.0 \\
    \midrule
    $+$ SFT   & 39.1  & 43.9 & 37.2 \\
    \rowcolor[HTML]{DAEFF9} $+$ \methodname  & 43.9 & 47.1 & 43.7 \\
    $\Delta$ & \hgreen{+3.5}  & \hgreen{+1.9} & \hgreen{+5.6}\\
    \bottomrule
    \end{tabular}%
  \label{tab:lisa_grounding}%
\end{table}%

\begin{table}[htbp]
\caption{
\small
\textbf{Open Vocabulary Object Detection Results on COCO dataset.} We trained on 65 base categories and tested on 15 novel categories.}
\vspace{-6mm}
\label{tab:coco_ov}
\begin{center}
\setlength{\tabcolsep}{10pt}
\scalebox{1.}
{
\begin{tabular}{l|llll}
\toprule
\multicolumn{1}{c}{\bf Models} &\multicolumn{1}{c}{\bf $mAP_{n}$}&\multicolumn{1}{c}{\bf $mAP_{b}$}&\multicolumn{1}{c}{\bf $mAP_{all}$}
\\
% \midrule
% Qwen2-VL-2B & 14.5 & 7.9 & 9.1\\
% \midrule
% $+$ SFT & 19.6 & 10.1 & 11.9\\
% \rowcolor[HTML]{DAEFF9} $+$ \methodname & 37.9 & 25.1 & 27.5\\
% $\Delta$ & \hgreen{+23.4}& \hgreen{+17.2}& \hgreen{+18.4}\\
% \midrule
% Qwen2-VL-7B & 31.0 & 19.6 & 21.7\\
% \midrule
% $+$ SFT   & 32.0 & 20.6 & 22.7\\
% \rowcolor[HTML]{DAEFF9} $+$ \methodname & 43.0 & 29.0 &31.6 \\
% $\Delta$ & \hgreen{+12.0}& \hgreen{+9.4}& \hgreen{+9.9}\\
\midrule
Qwen2-VL-2B & 9.8 & 6.0 & 6.7\\
\midrule
$+$ SFT & 13.6 & 7.8 & 8.9\\
\rowcolor[HTML]{DAEFF9} $+$ \methodname & 31.3 & 20.6 & 22.6\\
$\Delta$ & \hgreen{+21.5}& \hgreen{+14.6}& \hgreen{+15.9}\\
\midrule
Qwen2-VL-7B & 26.3 & 17.5 & 19.2\\
\midrule
$+$ SFT   & 25.7 & 17.5 & 19.0\\
\rowcolor[HTML]{DAEFF9} $+$ \methodname & 35.8 & 25.4 & 27.4\\
$\Delta$ & \hgreen{+9.5}& \hgreen{+7.9}& \hgreen{+8.2}\\
\bottomrule
\end{tabular}
}
\end{center}
\end{table}

\begin{table*}[t]
\caption{
\small
\textbf{Open Vocabulary Object Detection Results on LVIS dataset.} We trained on the 65 base categories of the COCO dataset and tested on the 13 rare categories of the LVIS dataset.}
\vspace{-6mm}
\label{tab:lvis_ov}
\begin{center}
\setlength{\tabcolsep}{4pt}
\scalebox{0.95}
{
\begin{tabular}{l|l|lllllllllllllll}
\toprule
\multicolumn{1}{c}{\bf Models}  &\multicolumn{1}{c}{\bf mAP} & \rotatebox{90}{casserole} & \rotatebox{90}{die} & \rotatebox{90}{egg roll} & \rotatebox{90}{futon}& \rotatebox{90}{garbage} & \rotatebox{90}{handsaw} & \rotatebox{90}{hippopotamus} & \rotatebox{90}{kitchen table} & \rotatebox{90}{mallet} & \rotatebox{90}{omelet} & \rotatebox{90}{shot glass}&\rotatebox{90}{stepladder}&\rotatebox{90} {sugar bowl}
\\
\midrule
GroudingDINO-B~\cite{liu2024grounding} & 23.9 & 17.1 & 0.0 & 2.4 & 47.5 & 27.7 & 13.4 & 15.2 & 92.5 & 0.0 & 26.6 & 16.0 & 41.0 & 10.7\\
\midrule
Qwen2-VL-2B & 2.7 & 1.6 & 1.2 & 0.0 & 2.4 & 0.0 & 10.0 & 0.0 & 13.4 & 0.2 & 4.7 & 2.1 & 0.0 & 0.0\\
\midrule
$+$ SFT  & 7.6 & 3.9 & 21.2 & 0.0 & 0.0 & 10.7 & 9.0 & 11.6 & 19.4 & 0.0 & 11.7 & 6.3 & 0.0 & 5.2\\
\rowcolor[HTML]{DAEFF9} $+$ \methodname & 20.7 & 24.5 & 23.4 & 2.0 & 16.0 & 27.7 & 20.2 & 14.4 & 45.8 & 11.1 & 22.7 & 6.0 & 6.0 & 40.2\\
$\Delta$ & \hgreen{+18.0} & \hgreen{+22.9} & \hgreen{+22.2} & \hgreen{+2.0} & \hgreen{+13.6} & \hgreen{+27.7} & \hgreen{+10.2} & \hgreen{+14.4} & \hgreen{+32.4} & \hgreen{+10.9} & \hgreen{+18.0} & \hgreen{+3.9} & \hgreen{+6.0} & \hgreen{+40.2}\\
\midrule
Qwen2-VL-7B  & 15.7 & 3.7 & 21.9 & 0.7 & 24.5 & 15.3 & 19.2 & 13.1 & 14.5 & 11.9 & 18.1 & 27.9 & 0.0 & 33.8\\
\midrule
$+$ SFT  & 24.0 & 20.8 & 25.4 & 0.6 & 41.8 & 12.2 & 19.2 & 18.8 & 42.5 & 11.9 & 15.3 & 27.9 & 28.1 & 47.8\\
\rowcolor[HTML]{DAEFF9} $+$ \methodname & 30.4 & 19.7 & 27.8 & 4.3 & 41.8 & 17.4 & 35.1 & 20.0 & 70.6 & 16.7 & 23.5 & 29.8 & 29.3 & 59.8\\
$\Delta$ & \hgreen{+14.7} & \hgreen{+16.0} & \hgreen{+5.9} & \hgreen{+3.6} & \hgreen{+17.3} & \hgreen{+2.1} & \hgreen{+15.9} & \hgreen{+6.9} & \hgreen{+56.1} & \hgreen{+4.8} & \hgreen{+5.4} & \hgreen{+1.9} & \hgreen{+29.3} & \hgreen{+26.0}\\
\bottomrule
\end{tabular}
}
\vspace{-4mm}
\end{center}
\end{table*}

\paragraph{Implementation Details} 
Our method is adaptable to various visual perception tasks. 
We employ a few-shot learning approach, providing the model with a minimal number of samples for training. For the image classification and object detection task, we adopt a few-shot setting to evaluate the model's fine-grained discriminative and recognition capability, applying reinforcement learning on limited data. 
Then, for the LISA~\cite{lai2024lisa} dataset focusing on reasoning grounding, which demands strong reasoning abilities, we train the model using \methodname and assess its reasoning and perception performance.
Lastly, for open-vocabulary object detection, we evaluate the model's generalization capability by training the Qwen2-VL-2/7B~\cite{wang2024qwen2} using \methodname on a subdivided COCO dataset containing 65 base classes. We then test it on 15 novel classes from COCO and 13 rare classes from LVIS~\cite{gupta2019lvis}. The model's visual perception and reasoning abilities are assessed in an open-vocabulary detection setting.
In our detection experiments, we first prompt the model to check whether the category is present in the image, then predict bound boxes for categories that exist in the images.

\subsection{Few-Shot Classification}
To demonstrate the extensive generalization ability of \methodname in the visual domain, we conduct few-shot experiments on fine-grained image classification. We selected four datasets: Flower102~\cite{flower102}, Pets37~\cite{pets37}, FGVC-Aircraft~\cite{fgvc}, and Car196~\cite{stanfordcars}, which contain dozens to hundreds of similar categories, adding significant difficulty to the classification task. 

As shown in Tab.~\ref{tab:cls}, with just one-shot of data, \methodname already delivers a significant performance boost (+24.3$\%$). In contrast, SFT shows a noticeable decline (-4.3$\%$) with the same minimal amount of data. Under the 4-shot setting, the performance of SFT is still slightly lower than the baseline, while the reinforcement fine-tuned model with \methodname achieves an average performance improvement of 25.9. Under the 8-shot and 16-shot settings, as the amount of data increases, SFT's performance slightly exceeds the baseline. However, SFT still lags significantly behind the performance of the \methodname. In Fig.\ref{fig:cls_case}, we present some inference cases of the model after reinforcement fine-tuning when handling fine-grained classification tasks. These results not only demonstrate the strong generalization ability of \methodname and its capacity to learn from limited data but also confirm that reinforcement fine-tuning, compared to SFT, leads to a genuine understanding of the task and deeper learning from reasoning. 

\subsection{Few-Shot Object Detection}
Few-shot learning has always been one of the core challenges faced by traditional visual models and large-scale vision-language models (LVLMs). Reinforcement fine-tuning provides a new solution to this problem by enabling the model to quickly learn and understand with a small amount of data. We selected eight categories from the COCO dataset, with 1, 2, 4, 8, and 16 images per category, to construct training sets with limited data. For the LVIS dataset, we select 6 rare categories. Since the training images for these rare categories are very sparse, with each category having between 1 and 10 images, we approximated this as a 10-shot setting. We then train the Qwen2-VL-2/7B model for 200 steps using both reinforcement fine-tuning and SFT, to evaluate the model’s learning ability with limited data.

As shown in Tab.~\ref{tab:few-shot_coco} and Tab.~\ref{tab:few-shot_lvis}, although both SFT and reinforcement fine-tuning can improve the model's recognition accuracy under the few-shot setting, the model after reinforcement fine-tuning consistently outperforms the SFT model by a large margin, maintaining a significant lead. On the COCO~\cite{coco} categories, as the training data increases, the SFT model reaches an average mAP of approximately $31$, while the reinforcement fine-tuned model approaches $47$. In the LVIS~\cite{gupta2019lvis} few-shot experimental results shown in Tab.~\ref{tab:few-shot_lvis}, for the six more challenging rare categories in LVIS, reinforcement fine-tuning still outperforms SFT. The results in Tab.~\ref{tab:few-shot_coco} and Tab.~\ref{tab:few-shot_lvis} clearly demonstrate the exceptional performance of reinforcement fine-tuning in the few-shot setting, where the model achieves a significant improvement in visual perception capabilities through reinforcement learning with only a small amount of data.

We further test on some abstract out-of-domain datasets. We selected the MG (Monster Girls) dataset, which contains different types of anime-style monster girls. By using out-of-domain data, we increased the difficulty of both model recognition and reasoning, and conducted experiments under 4-shot and 16-shot settings. The results, shown in Tab.~\ref{tab:mg_result}, indicate that reinforcement fine-tuning achieved a significant performance improvement, surpassing supervised fine-tuning (SFT).

\subsection{Reasoning Grounding}
Another crucial aspect of vision-language intelligence is grounding the exact object according to user needs. Previous specialized detection systems lack reasoning abilities and fail to fully understand the user's intentions. Pioneered by LISA~\cite{lai2024lisa}, there have been works done to enable large language models (LLMs) to output control tokens for other models (such as SAM~\cite{kirillov2023segment}) or directly predict bounding box coordinates~\cite{wang2024qwen2,rasheed2024glamm} through supervised fine-tuning. In our work, we explore the use of \methodname in this task and find that reinforcement learning (RL) leads to significant improvements over supervised fine-tuning.

We finetune Qwen2-VL 2B/7B model~\cite{wang2024qwen2} using \methodname and supervised fine-tuning (SFT) on the LISA training set, which consists of 239 images with reasoning grounding objects. We follow the same test setting with LISA and compare the results of SFT and our method, both with 500 fine-tuning steps. As shown in \cref{tab:lisa_grounding}, \methodname significantly improves the final results in terms of bounding box IoU compared to SFT. Additionally, we prompt SAM~\cite{kirillov2023segment} with the Qwen2-VL predicted bounding box to generate the segmentation mask (evaluated using gIoU). \methodname significantly enhances grounding ability and outperforms previous specialized detection systems. Qualitative results are visualized in Fig.~\ref{fig:lisa_case}, where the thinking process significantly improves the ability to reason and grounding accuracy. Through \methodname, Qwen2-VL learns to think critically and carefully examine the image to produce accurate grounding results.

\subsection{Open Vocabulary Object Detection}
The advantage of \methodname over SFT arises from the former’s true deep understanding of the task, rather than merely memorizing the data. To further demonstrate the powerful generalization ability of reinforcement fine-tuning, we conduct open vocabulary object detection experiments. We first randomly sampled 6K annotations from the COCO dataset, which included 65 base categories. We perform \methodname and SFT on the Qwen2-VL-2/7B model~\cite{wang2024qwen2} using this data, and test the model on 15 new categories it has never seen before. To increase the difficulty, we further test 13 rare categories from the LVIS~\cite{gupta2019lvis} dataset.

As shown in Tab.~\ref{tab:coco_ov} and Tab.~\ref{tab:lvis_ov}, after reinforcement fine-tuning, the Qwen2-VL-2/7B model achieves an average mAP increase of $21.5$ and $9.5$ on 15 new categories from the COCO dataset. On the more challenging rare categories of the LVIS~\cite{gupta2019lvis} dataset, mAP increased by $18.0$ and $14.7$. The \methodname not only transfers its detection capabilities from the COCO base categories to the new COCO categories but also achieves significant improvements on the more challenging rare categories of LVIS. Notably, for some rare LVIS categories in Tab.~\ref{tab:lvis_ov}, the original or SFT-trained models cannot recognize these categories, resulting in 0 AP. However, after reinforcement fine-tuning, the model shows a qualitative leap from 0 to 1 in recognizing these previously unidentifiable categories (such as egg roll and futon). This demonstrates that \methodname has a significant impact on improving the performance and generalization ability in visual recognition for LVLMs.

\label{sec_4_experiments}

\section{Conclusion}
In this paper, we introduce Visual Reinforcement Fine-tuning (\methodname), the first approach to adapt the GRPO-based reinforcement learning strategy for enhancing the visual perception and grounding ability of LVLMs. By using a rule-based verifiable reward system, \methodname reduces the need for manual labeling and simplifies reward computation, achieving significant improvements across various visual perception tasks. Extensive experiments show that \methodname excels in fine-grained classification, open vocabulary detection, reasoning grounding and few-shot learning tasks. It outperforms supervised fine-tuning (SFT) with minimal data and shows strong generalization. This work demonstrates the potential of reinforcement learning to enhance the capabilities of LVLMs, making them more efficient and effective in visual perception tasks. 
\label{sec_5_conclusion}

\newpage
{
    \small
    \bibliographystyle{ieeenat_fullname}
    \bibliography{main}

\begin{thebibliography}{54}
\providecommand{\natexlab}[1]{#1}
\providecommand{\url}[1]{\texttt{#1}}
\expandafter\ifx\csname urlstyle\endcsname\relax
  \providecommand{\doi}[1]{doi: #1}\else
  \providecommand{\doi}{doi: \begingroup \urlstyle{rm}\Url}\fi

\bibitem[Abdulhai et~al.(2023)Abdulhai, White, Snell, Sun, Hong, Zhai, Xu, and Levine]{LMRL-Gym}
Marwa Abdulhai, Isadora White, Charlie Snell, Charles Sun, Joey Hong, Yuexiang Zhai, Kelvin Xu, and Sergey Levine.
\newblock Lmrl gym: Benchmarks for multi-turn reinforcement learning with language models.
\newblock \emph{arXiv preprint arXiv:2311.18232}, 2023.

\bibitem[Cai et~al.(2024)Cai, Cao, Chen, Chen, Chen, Chen, Chen, Chen, Chen, Chu, et~al.]{cai2024internlm2}
Zheng Cai, Maosong Cao, Haojiong Chen, Kai Chen, Keyu Chen, Xin Chen, Xun Chen, Zehui Chen, Zhi Chen, Pei Chu, et~al.
\newblock Internlm2 technical report.
\newblock \emph{arXiv preprint arXiv:2403.17297}, 2024.

\bibitem[Carta et~al.(2023)Carta, Romac, Wolf, Lamprier, Sigaud, and Oudeyer]{Grounding-large-language-models}
Thomas Carta, Cl{\'e}ment Romac, Thomas Wolf, Sylvain Lamprier, Olivier Sigaud, and Pierre-Yves Oudeyer.
\newblock Grounding large language models in interactive environments with online reinforcement learning.
\newblock In \emph{ICLR}, 2023.

\bibitem[Guo et~al.(2025)Guo, Yang, Zhang, Song, Zhang, Xu, Zhu, Ma, Wang, Bi, et~al.]{DeepSeek-R1}
Daya Guo, Dejian Yang, Haowei Zhang, Junxiao Song, Ruoyu Zhang, Runxin Xu, Qihao Zhu, Shirong Ma, Peiyi Wang, Xiao Bi, et~al.
\newblock Deepseek-r1: Incentivizing reasoning capability in llms via reinforcement learning.
\newblock \emph{arXiv preprint arXiv:2501.12948}, 2025.

\bibitem[Gupta et~al.(2019)Gupta, Dollar, and Girshick]{gupta2019lvis}
Agrim Gupta, Piotr Dollar, and Ross Girshick.
\newblock Lvis: A dataset for large vocabulary instance segmentation.
\newblock In \emph{Proceedings of the IEEE/CVF conference on computer vision and pattern recognition}, pages 5356--5364, 2019.

\bibitem[Hui et~al.(2024)Hui, Yang, Cui, Yang, Liu, Zhang, Liu, Zhang, Yu, Lu, et~al.]{hui2024qwen2coder}
Binyuan Hui, Jian Yang, Zeyu Cui, Jiaxi Yang, Dayiheng Liu, Lei Zhang, Tianyu Liu, Jiajun Zhang, Bowen Yu, Keming Lu, et~al.
\newblock Qwen2. 5-coder technical report.
\newblock \emph{arXiv preprint arXiv:2409.12186}, 2024.

\bibitem[Jaech et~al.(2024)Jaech, Kalai, Lerer, Richardson, El-Kishky, Low, Helyar, Madry, Beutel, Carney, et~al.]{OpenAI_O1}
Aaron Jaech, Adam Kalai, Adam Lerer, Adam Richardson, Ahmed El-Kishky, Aiden Low, Alec Helyar, Aleksander Madry, Alex Beutel, Alex Carney, et~al.
\newblock Openai o1 system card.
\newblock \emph{arXiv:2412.16720}, 2024.

\bibitem[Jiao et~al.(2024)Jiao, Guo, Zhang, Chen, Joty, and Wei]{jiao2024preferencecode}
Fangkai Jiao, Geyang Guo, Xingxing Zhang, Nancy~F Chen, Shafiq Joty, and Furu Wei.
\newblock Preference optimization for reasoning with pseudo feedback.
\newblock \emph{arXiv preprint arXiv:2411.16345}, 2024.

\bibitem[Kirillov et~al.(2023)Kirillov, Mintun, Ravi, Mao, Rolland, Gustafson, Xiao, Whitehead, Berg, Lo, et~al.]{kirillov2023segment}
Alexander Kirillov, Eric Mintun, Nikhila Ravi, Hanzi Mao, Chloe Rolland, Laura Gustafson, Tete Xiao, Spencer Whitehead, Alexander~C Berg, Wan-Yen Lo, et~al.
\newblock Segment anything.
\newblock In \emph{Proceedings of the IEEE/CVF international conference on computer vision}, pages 4015--4026, 2023.

\bibitem[Krause et~al.(2013)Krause, Stark, Deng, and Fei-Fei]{stanfordcars}
Jonathan Krause, Michael Stark, Jia Deng, and Li Fei-Fei.
\newblock 3d object representations for fine-grained categorization.
\newblock In \emph{ICCV workshops}, 2013.

\bibitem[Lai et~al.(2024)Lai, Tian, Chen, Li, Yuan, Liu, and Jia]{lai2024lisa}
Xin Lai, Zhuotao Tian, Yukang Chen, Yanwei Li, Yuhui Yuan, Shu Liu, and Jiaya Jia.
\newblock Lisa: Reasoning segmentation via large language model.
\newblock In \emph{Proceedings of the IEEE/CVF Conference on Computer Vision and Pattern Recognition}, pages 9579--9589, 2024.

\bibitem[Lambert et~al.(2024)Lambert, Morrison, Pyatkin, Huang, Ivison, Brahman, Miranda, Liu, Dziri, Lyu, et~al.]{lambert2024t}
Nathan Lambert, Jacob Morrison, Valentina Pyatkin, Shengyi Huang, Hamish Ivison, Faeze Brahman, Lester James~V Miranda, Alisa Liu, Nouha Dziri, Shane Lyu, et~al.
\newblock T$\backslash$" ulu 3: Pushing frontiers in open language model post-training.
\newblock \emph{arXiv preprint arXiv:2411.15124}, 2024.

\bibitem[Li et~al.(2024)Li, Zhang, Guo, Zhang, Li, Zhang, Zhang, Zhang, Li, Liu, et~al.]{li2024llavaov}
Bo Li, Yuanhan Zhang, Dong Guo, Renrui Zhang, Feng Li, Hao Zhang, Kaichen Zhang, Peiyuan Zhang, Yanwei Li, Ziwei Liu, et~al.
\newblock Llava-onevision: Easy visual task transfer.
\newblock \emph{arXiv preprint arXiv:2408.03326}, 2024.

\bibitem[Liang et~al.(2023)Liang, Wu, Dai, Li, Zhao, Zhang, Zhang, Vajda, and Marculescu]{liang2023open}
Feng Liang, Bichen Wu, Xiaoliang Dai, Kunpeng Li, Yinan Zhao, Hang Zhang, Peizhao Zhang, Peter Vajda, and Diana Marculescu.
\newblock Open-vocabulary semantic segmentation with mask-adapted clip.
\newblock In \emph{Proceedings of the IEEE/CVF conference on computer vision and pattern recognition}, pages 7061--7070, 2023.

\bibitem[Lin et~al.(2014)Lin, Maire, Belongie, Hays, Perona, Ramanan, Doll{\'a}r, and Zitnick]{coco}
Tsung-Yi Lin, Michael Maire, Serge Belongie, James Hays, Pietro Perona, Deva Ramanan, Piotr Doll{\'a}r, and C~Lawrence Zitnick.
\newblock Microsoft coco: Common objects in context.
\newblock In \emph{Computer Vision--ECCV 2014: 13th European Conference, Zurich, Switzerland, September 6-12, 2014, Proceedings, Part V 13}, pages 740--755. Springer, 2014.

\bibitem[Liu et~al.(2024{\natexlab{a}})Liu, Feng, Xue, Wang, Wu, Lu, Zhao, Deng, Zhang, Ruan, et~al.]{liu2024deepseekv3}
Aixin Liu, Bei Feng, Bing Xue, Bingxuan Wang, Bochao Wu, Chengda Lu, Chenggang Zhao, Chengqi Deng, Chenyu Zhang, Chong Ruan, et~al.
\newblock Deepseek-v3 technical report.
\newblock \emph{arXiv preprint arXiv:2412.19437}, 2024{\natexlab{a}}.

\bibitem[Liu et~al.(2024{\natexlab{b}})Liu, Zeng, Liu, Yan, He, Wang, Yan, Liu, and Zhou]{liu2024skywork}
Chris~Yuhao Liu, Liang Zeng, Jiacai Liu, Rui Yan, Jujie He, Chaojie Wang, Shuicheng Yan, Yang Liu, and Yahui Zhou.
\newblock {Skywork-Reward}: Bag of tricks for reward modeling in llms.
\newblock \emph{arXiv preprint arXiv:2410.18451}, 2024{\natexlab{b}}.

\bibitem[Liu et~al.(2024{\natexlab{c}})Liu, Zeng, Ren, Li, Zhang, Yang, Jiang, Li, Yang, Su, et~al.]{liu2024grounding}
Shilong Liu, Zhaoyang Zeng, Tianhe Ren, Feng Li, Hao Zhang, Jie Yang, Qing Jiang, Chunyuan Li, Jianwei Yang, Hang Su, et~al.
\newblock Grounding dino: Marrying dino with grounded pre-training for open-set object detection.
\newblock In \emph{European Conference on Computer Vision}, pages 38--55. Springer, 2024{\natexlab{c}}.

\bibitem[Liu et~al.(2024{\natexlab{d}})Liu, Zang, Dong, Zhang, Cao, Duan, He, Xiong, Lin, and Wang]{liu2024mia}
Ziyu Liu, Yuhang Zang, Xiaoyi Dong, Pan Zhang, Yuhang Cao, Haodong Duan, Conghui He, Yuanjun Xiong, Dahua Lin, and Jiaqi Wang.
\newblock Mia-dpo: Multi-image augmented direct preference optimization for large vision-language models.
\newblock \emph{arXiv preprint arXiv:2410.17637}, 2024{\natexlab{d}}.

\bibitem[Luong et~al.(2024)Luong, Zhang, Jie, Sun, Jin, and Li]{luong2024reft}
Trung~Quoc Luong, Xinbo Zhang, Zhanming Jie, Peng Sun, Xiaoran Jin, and Hang Li.
\newblock Reft: Reasoning with reinforced fine-tuning, 2024.

\bibitem[Maji et~al.(2013)Maji, Rahtu, Kannala, Blaschko, and Vedaldi]{fgvc}
Subhransu Maji, Esa Rahtu, Juho Kannala, Matthew Blaschko, and Andrea Vedaldi.
\newblock Fine-grained visual classification of aircraft.
\newblock \emph{arXiv preprint arXiv:1306.5151}, 2013.

\bibitem[Nilsback and Zisserman(2008)]{flower102}
Maria-Elena Nilsback and Andrew Zisserman.
\newblock Automated flower classification over a large number of classes.
\newblock In \emph{ICVGIP}, 2008.

\bibitem[OpenAI(2024)]{2024gpt4o}
OpenAI.
\newblock Hello gpt-4o, 2024.

\bibitem[OpenAI(2025)]{OpenAI_O3}
OpenAI.
\newblock Openai o3-mini system card, 2025.

\bibitem[Ouyang et~al.(2022{\natexlab{a}})Ouyang, Wu, Jiang, Almeida, Wainwright, Mishkin, Zhang, Agarwal, Slama, Ray, et~al.]{Training-language-models}
Long Ouyang, Jeffrey Wu, Xu Jiang, Diogo Almeida, Carroll Wainwright, Pamela Mishkin, Chong Zhang, Sandhini Agarwal, Katarina Slama, Alex Ray, et~al.
\newblock Training language models to follow instructions with human feedback.
\newblock In \emph{NeurIPS}, 2022{\natexlab{a}}.

\bibitem[Ouyang et~al.(2022{\natexlab{b}})Ouyang, Wu, Jiang, Almeida, Wainwright, Mishkin, Zhang, Agarwal, Slama, Ray, Schulman, Hilton, Kelton, Miller, Simens, Askell, Welinder, Christiano, Leike, and Lowe]{ouyang2022training}
Long Ouyang, Jeff Wu, Xu Jiang, Diogo Almeida, Carroll~L. Wainwright, Pamela Mishkin, Chong Zhang, Sandhini Agarwal, Katarina Slama, Alex Ray, John Schulman, Jacob Hilton, Fraser Kelton, Luke~E. Miller, Maddie Simens, Amanda Askell, Peter Welinder, Paul~Francis Christiano, Jan Leike, and Ryan~J. Lowe.
\newblock Training language models to follow instructions with human feedback.
\newblock In \emph{NeurIPS}, 2022{\natexlab{b}}.

\bibitem[Parkhi et~al.(2012)Parkhi, Vedaldi, Zisserman, and Jawahar]{pets37}
Omkar~M Parkhi, Andrea Vedaldi, Andrew Zisserman, and CV Jawahar.
\newblock Cats and dogs.
\newblock In \emph{CVPR}, 2012.

\bibitem[Ramamurthy et~al.(2023)Ramamurthy, Ammanabrolu, Brantley, Hessel, Sifa, Bauckhage, Hajishirzi, and Choi]{RL4LMs}
Rajkumar Ramamurthy, Prithviraj Ammanabrolu, Kiant{\'e} Brantley, Jack Hessel, Rafet Sifa, Christian Bauckhage, Hannaneh Hajishirzi, and Yejin Choi.
\newblock Is reinforcement learning (not) for natural language processing: Benchmarks, baselines, and building blocks for natural language policy optimization.
\newblock In \emph{ICLR}, 2023.

\bibitem[Rasheed et~al.(2024)Rasheed, Maaz, Shaji, Shaker, Khan, Cholakkal, Anwer, Xing, Yang, and Khan]{rasheed2024glamm}
Hanoona Rasheed, Muhammad Maaz, Sahal Shaji, Abdelrahman Shaker, Salman Khan, Hisham Cholakkal, Rao~M Anwer, Eric Xing, Ming-Hsuan Yang, and Fahad~S Khan.
\newblock Glamm: Pixel grounding large multimodal model.
\newblock In \emph{Proceedings of the IEEE/CVF Conference on Computer Vision and Pattern Recognition}, pages 13009--13018, 2024.

\bibitem[Schulman et~al.(2017)Schulman, Wolski, Dhariwal, Radford, and Klimov]{PPO}
John Schulman, Filip Wolski, Prafulla Dhariwal, Alec Radford, and Oleg Klimov.
\newblock Proximal policy optimization algorithms.
\newblock \emph{arXiv:1707.06347}, 2017.

\bibitem[Shao et~al.(2024)Shao, Wang, Zhu, Xu, Song, Bi, Zhang, Zhang, Li, Wu, et~al.]{grpo}
Zhihong Shao, Peiyi Wang, Qihao Zhu, Runxin Xu, Junxiao Song, Xiao Bi, Haowei Zhang, Mingchuan Zhang, YK Li, Y Wu, et~al.
\newblock Deepseekmath: Pushing the limits of mathematical reasoning in open language models.
\newblock \emph{arXiv preprint arXiv:2402.03300}, 2024.

\bibitem[Snell et~al.(2023)Snell, Kostrikov, Su, Yang, and Levine]{ILQL}
Charlie~Victor Snell, Ilya Kostrikov, Yi Su, Sherry Yang, and Sergey Levine.
\newblock Offline {RL} for natural language generation with implicit language q learning.
\newblock In \emph{ICLR}, 2023.

\bibitem[Stiennon et~al.(2022)Stiennon, Ouyang, Wu, Ziegler, Lowe, Voss, Radford, Amodei, and Christiano]{Learning-to-summarize}
Nisan Stiennon, Long Ouyang, Jeffrey Wu, Daniel Ziegler, Ryan Lowe, Chelsea Voss, Alec Radford, Dario Amodei, and Paul~F Christiano.
\newblock Learning to summarize with human feedback.
\newblock In \emph{NeurIPS}, 2022.

\bibitem[Sun et~al.(2023{\natexlab{a}})Sun, Shen, Cao, Liu, Li, Shen, Gan, Gui, Wang, Yang, et~al.]{llavarlhf}
Zhiqing Sun, Sheng Shen, Shengcao Cao, Haotian Liu, Chunyuan Li, Yikang Shen, Chuang Gan, Liang-Yan Gui, Yu-Xiong Wang, Yiming Yang, et~al.
\newblock Aligning large multimodal models with factually augmented rlhf.
\newblock \emph{arXiv preprint arXiv:2309.14525}, 2023{\natexlab{a}}.

\bibitem[Sun et~al.(2023{\natexlab{b}})Sun, Shen, Cao, Liu, Li, Shen, Gan, Gui, Wang, Yang, et~al.]{sun2023aligning}
Zhiqing Sun, Sheng Shen, Shengcao Cao, Haotian Liu, Chunyuan Li, Yikang Shen, Chuang Gan, Liang-Yan Gui, Yu-Xiong Wang, Yiming Yang, et~al.
\newblock Aligning large multimodal models with factually augmented rlhf.
\newblock \emph{arXiv preprint arXiv:2309.14525}, 2023{\natexlab{b}}.

\bibitem[Sun et~al.(2024)Sun, Shen, Cao, Liu, Li, Shen, Gan, Gui, Wang, Yang, et~al.]{Aligning-LLMs-with-RLHF}
Zhiqing Sun, Sheng Shen, Shengcao Cao, Haotian Liu, Chunyuan Li, Yikang Shen, Chuang Gan, Liang-Yan Gui, Yu-Xiong Wang, Yiming Yang, et~al.
\newblock Aligning large multimodal models with factually augmented rlhf.
\newblock In \emph{ACL}, 2024.

\bibitem[Team et~al.(2025)Team, Du, Gao, Xing, Jiang, Chen, Li, Xiao, Du, Liao, et~al.]{team2025kimi}
Kimi Team, Angang Du, Bofei Gao, Bowei Xing, Changjiu Jiang, Cheng Chen, Cheng Li, Chenjun Xiao, Chenzhuang Du, Chonghua Liao, et~al.
\newblock Kimi k1. 5: Scaling reinforcement learning with llms.
\newblock \emph{arXiv preprint arXiv:2501.12599}, 2025.

\bibitem[Wang et~al.(2024)Wang, Bai, Tan, Wang, Fan, Bai, Chen, Liu, Wang, Ge, et~al.]{wang2024qwen2}
Peng Wang, Shuai Bai, Sinan Tan, Shijie Wang, Zhihao Fan, Jinze Bai, Keqin Chen, Xuejing Liu, Jialin Wang, Wenbin Ge, et~al.
\newblock Qwen2-vl: Enhancing vision-language model's perception of the world at any resolution.
\newblock \emph{arXiv preprint arXiv:2409.12191}, 2024.

\bibitem[Yang et~al.(2024)Yang, Zhang, Hui, Gao, Yu, Li, Liu, Tu, Zhou, Lin, et~al.]{yang2024qwen2math}
An Yang, Beichen Zhang, Binyuan Hui, Bofei Gao, Bowen Yu, Chengpeng Li, Dayiheng Liu, Jianhong Tu, Jingren Zhou, Junyang Lin, et~al.
\newblock Qwen2. 5-math technical report: Toward mathematical expert model via self-improvement.
\newblock \emph{arXiv preprint arXiv:2409.12122}, 2024.

\bibitem[Yao et~al.(2023)Yao, Zhao, Yu, Du, Shafran, Narasimhan, and Cao]{ReAct}
Shunyu Yao, Jeffrey Zhao, Dian Yu, Nan Du, Izhak Shafran, Karthik~R Narasimhan, and Yuan Cao.
\newblock React: Synergizing reasoning and acting in language models.
\newblock In \emph{ICLR}, 2023.

\bibitem[Ying et~al.(2024)Ying, Zhang, Li, Zhou, Shao, Fei, Ma, Hong, Liu, Wang, et~al.]{ying2024internlmmath}
Huaiyuan Ying, Shuo Zhang, Linyang Li, Zhejian Zhou, Yunfan Shao, Zhaoye Fei, Yichuan Ma, Jiawei Hong, Kuikun Liu, Ziyi Wang, et~al.
\newblock Internlm-math: Open math large language models toward verifiable reasoning.
\newblock \emph{arXiv preprint arXiv:2402.06332}, 2024.

\bibitem[Yu et~al.(2024{\natexlab{a}})Yu, Yao, Zhang, He, Han, Cui, Hu, Liu, Zheng, Sun, et~al.]{yu2024rlhfv}
Tianyu Yu, Yuan Yao, Haoye Zhang, Taiwen He, Yifeng Han, Ganqu Cui, Jinyi Hu, Zhiyuan Liu, Hai-Tao Zheng, Maosong Sun, et~al.
\newblock {RlHF-V}: Towards trustworthy mllms via behavior alignment from fine-grained correctional human feedback.
\newblock In \emph{CVPR}, 2024{\natexlab{a}}.

\bibitem[Yu et~al.(2024{\natexlab{b}})Yu, Zhang, Yao, Dang, Chen, Lu, Cui, He, Liu, Chua, et~al.]{yu2024rlaif}
Tianyu Yu, Haoye Zhang, Yuan Yao, Yunkai Dang, Da Chen, Xiaoman Lu, Ganqu Cui, Taiwen He, Zhiyuan Liu, Tat-Seng Chua, et~al.
\newblock {RLAIF-V}: Aligning mllms through open-source ai feedback for super gpt-4v trustworthiness.
\newblock \emph{arXiv preprint arXiv:2405.17220}, 2024{\natexlab{b}}.

\bibitem[Zang et~al.(2024)Zang, Li, Han, Zhou, and Loy]{zang2024contextual}
Yuhang Zang, Wei Li, Jun Han, Kaiyang Zhou, and Chen~Change Loy.
\newblock Contextual object detection with multimodal large language models.
\newblock \emph{IJCV}, 2024.

\bibitem[Zang et~al.(2025)Zang, Dong, Zhang, Cao, Liu, Ding, Wu, Ma, Duan, Zhang, et~al.]{zang2025internlm}
Yuhang Zang, Xiaoyi Dong, Pan Zhang, Yuhang Cao, Ziyu Liu, Shengyuan Ding, Shenxi Wu, Yubo Ma, Haodong Duan, Wenwei Zhang, et~al.
\newblock A simple yet effective multi-modal reward model.
\newblock \emph{arXiv preprint arXiv:2501.12368}, 2025.

\bibitem[Zhang et~al.(2024{\natexlab{a}})Zhang, Li, Dong, Xu, Zhang, Su, Liu, and Jin]{zhang2024codedpo}
Kechi Zhang, Ge Li, Yihong Dong, Jingjing Xu, Jun Zhang, Jing Su, Yongfei Liu, and Zhi Jin.
\newblock Codedpo: Aligning code models with self generated and verified source code.
\newblock \emph{arXiv preprint arXiv:2410.05605}, 2024{\natexlab{a}}.

\bibitem[Zhang et~al.(2024{\natexlab{b}})Zhang, Dong, Zang, Cao, Qian, Chen, Guo, Duan, Wang, Ouyang, et~al.]{zhang2024internlm}
Pan Zhang, Xiaoyi Dong, Yuhang Zang, Yuhang Cao, Rui Qian, Lin Chen, Qipeng Guo, Haodong Duan, Bin Wang, Linke Ouyang, et~al.
\newblock Internlm-xcomposer-2.5: A versatile large vision language model supporting long-contextual input and output.
\newblock \emph{arXiv preprint arXiv:2407.03320}, 2024{\natexlab{b}}.

\bibitem[Zhang et~al.(2024{\natexlab{c}})Zhang, Wu, Yang, Shu, Xiao, Kong, and Sang]{zhang2024o1}
Yuxiang Zhang, Shangxi Wu, Yuqi Yang, Jiangming Shu, Jinlin Xiao, Chao Kong, and Jitao Sang.
\newblock o1-coder: an o1 replication for coding.
\newblock \emph{arXiv preprint arXiv:2412.00154}, 2024{\natexlab{c}}.

\bibitem[Zhao et~al.(2023)Zhao, Wang, Ouyang, Dong, Wang, and He]{hadpo}
Zhiyuan Zhao, Bin Wang, Linke Ouyang, Xiaoyi Dong, Jiaqi Wang, and Conghui He.
\newblock Beyond hallucinations: Enhancing lvlms through hallucination-aware direct preference optimization.
\newblock \emph{arXiv preprint arXiv:2311.16839}, 2023.

\bibitem[Zhou et~al.(2024{\natexlab{a}})Zhou, Cui, Rafailov, Finn, and Yao]{povid}
Yiyang Zhou, Chenhang Cui, Rafael Rafailov, Chelsea Finn, and Huaxiu Yao.
\newblock Aligning modalities in vision large language models via preference fine-tuning.
\newblock \emph{arXiv preprint arXiv:2402.11411}, 2024{\natexlab{a}}.

\bibitem[Zhou et~al.(2024{\natexlab{b}})Zhou, Cui, Rafailov, Finn, and Yao]{zhou2024aligning}
Yiyang Zhou, Chenhang Cui, Rafael Rafailov, Chelsea Finn, and Huaxiu Yao.
\newblock Aligning modalities in vision large language models via preference fine-tuning.
\newblock \emph{arXiv preprint arXiv:2402.11411}, 2024{\natexlab{b}}.

\bibitem[Zhou et~al.(2024{\natexlab{c}})Zhou, Zanette, Pan, Levine, and Kumar]{ArCHer}
Yifei Zhou, Andrea Zanette, Jiayi Pan, Sergey Levine, and Aviral Kumar.
\newblock Archer: Training language model agents via hierarchical multi-turn rl.
\newblock In \emph{ICML}, 2024{\natexlab{c}}.

\bibitem[Ziegler et~al.(2019)Ziegler, Stiennon, Wu, Brown, Radford, Amodei, Christiano, and Irving]{lm-human-preferences}
Daniel~M Ziegler, Nisan Stiennon, Jeffrey Wu, Tom~B Brown, Alec Radford, Dario Amodei, Paul Christiano, and Geoffrey Irving.
\newblock Fine-tuning language models from human preferences.
\newblock \emph{arXiv:1909.08593}, 2019.

\bibitem[Zou et~al.(2023)Zou, Dou, Yang, Gan, Li, Li, Dai, Behl, Wang, Yuan, et~al.]{zou2023generalized}
Xueyan Zou, Zi-Yi Dou, Jianwei Yang, Zhe Gan, Linjie Li, Chunyuan Li, Xiyang Dai, Harkirat Behl, Jianfeng Wang, Lu Yuan, et~al.
\newblock Generalized decoding for pixel, image, and language.
\newblock In \emph{Proceedings of the IEEE/CVF conference on computer vision and pattern recognition}, pages 15116--15127, 2023.

\end{thebibliography}
}

% \newpage
\appendix
% \onecolumn

% \section{Appendix}
% \input{sec/6_appendix.tex}
% \label{sec_6_appendix}

\end{CJK}
\end{document}